\documentclass[conference]{IEEEtran}

\usepackage{amsmath,amssymb,amsfonts}
\usepackage{algorithm}
\usepackage{algorithmic}
\usepackage{graphicx}
\usepackage{multicol}
\usepackage{xcolor}
\usepackage{tabularx}
\usepackage{float}
\usepackage{color, colortbl}
\usepackage{graphicx}
\usepackage{amsmath}
\usepackage{amssymb}
\usepackage{booktabs}
\usepackage{multirow}
\usepackage{multicol}
\usepackage{xcolor}
\usepackage{tabularx}
\usepackage{float}
\usepackage{hyperref}
\usepackage{setspace}
\usepackage[numbers]{natbib} 
\usepackage[normalem]{ulem}

\definecolor{Cyan}{rgb}{0.88,1,1}

\begin{document}

\title{USTM: Unified Spatial and Temporal Modeling for Continuous Sign Language Recognition}

\author{\IEEEauthorblockN{Ahmed Abul Hasanaath, Hamzah Luqman}
\IEEEauthorblockA{\textit{College of Information and Computer Science} \\
\textit{King Fahd University of Petroleum and Minerals, Khobar, KSA}\\}
}
\maketitle


\begin{abstract}
Continuous sign language recognition (CSLR) requires precise spatio-temporal modeling to accurately recognize sequences of gestures in videos. Existing frameworks often rely on CNN-based spatial backbones combined with temporal convolution or recurrent modules. These techniques fail in capturing fine-grained hand and facial cues and modeling long-range temporal dependencies. To address these limitations, we propose the Unified Spatio-Temporal Modeling (USTM) framework, a spatio-temporal encoder that effectively models complex patterns using a combination of a Swin Transformer backbone enhanced with lightweight temporal adapter with positional embeddings (TAPE). Our framework captures fine-grained spatial features alongside short and long-term temporal context, enabling robust sign language recognition from RGB videos without relying on multi-stream inputs or auxiliary modalities. Extensive experiments on benchmarked datasets including PHOENIX14, PHOENIX14T, and CSL-Daily demonstrate that USTM achieves state-of-the-art performance against RGB-based as well as multi-modal CSLR approaches, while maintaining competitive performance against multi-stream approaches. These results highlight the strength and efficacy of the USTM framework for CSLR. The code is available at \url{https://github.com/gufranSabri/USTM}

\end{abstract}


\section{Introduction}

Sign language is a fully-fledged visual language used by deaf communities for communication. This language is characterized by a complex combination of hand motion, facial expression, and body posture \cite{wadhawan2021sign}. Unlike spoken languages, which rely on sequential acoustic signals, sign languages are inherently multimodal and spatio-temporal, which encode information simultaneously through both manual and non-manual cues \cite{sidig2021karsl}. Sign Language Recognition (SLR) aims to automatically recognize sign language gestures in videos and translate them into written or spoken language. SLR is important serving as a crucial step toward integrating deaf people in the community. It can be broadly categorized based on the recognized sign unit into isolated SLR (ISLR) and continuous SLR (CSLR). ISLR systems focus on recognizing individual signs that are already segmented, whereas CSLR systems aim to recognize sign language sentences in which signs appear fluidly in sequence, much like natural speech.
CSLR is more challenging than ISLR due to the lack of temporal information that marks the precise start and end of each sign in the sentence making it a weakly supervised learning problem \cite{luqman2023arabsign}. This ambiguity in temporal boundaries, coupled with high inter-signer and intra-signer variability, makes CSLR more challenging than its isolated counterpart.

Several approaches have been proposed in the literature for CSLR. These approaches can be categorized into single-modality and multi-modality methods.
The majority of recent CSLR studies \cite{cheng2020fully, min2021visual, hao2021self, xie2023multi, hu2023self, cui2023spatial, alyami2025swin} predominantly utilize RGB-based modality. Cheng et al.~\cite{cheng2020fully} proposed a fully convolutional network with gloss feature enhancement to improve alignment between visual features and gloss sequences. To better capture local motion cues, Xie et al.~\cite{xie2023multi} introduced a multi-scale temporal similarity fusion network with position-aware convolutions. VAC loss~\cite{min2021visual} further improved alignment by enforcing visual consistency between encoder and decoder features. Attention and distillation-based methods, such as SEN~\cite{hu2023self} and SMKD~\cite{hao2021self}, emphasized key regions (e.g., hands and face) and reduced prediction spikes. Temporal Lift Pooling~\cite{hu2022temporal} and CorrNet~\cite{hu2023continuous} techniques tackled temporal downsampling and cross-frame correlation modeling. To address scalability on longer videos, transformer-based models, such as ST-Transformer~\cite{cui2023spatial} and Swin-MSTP~\cite{alyami2025swin}, leveraged frame chunking and multiscale temporal modules, showing strong performance with RGB-only input.

Multi-modal and multi-stream approaches have also shown a strong promise in advancing CSLR by incorporating complementary information beyond raw RGB frames. Multi-modal models leverage additional cues, such as pose, hand crops, and optical flow, to enrich spatial and temporal representations. SignBERT~\cite{zhou2021signbert} incorporates hand-focused visual cues into a transformer backbone to enhance signer generalization, while C2SLR~\cite{zuo2022c2slr} employs keypoint-guided attention alongside semantic constraints. STMC~\cite{zhou2021spatial} and Deep Neural Fusion~\cite{cui2019deep} combine pose, RGB, and optical flow modalities to enhance temporal modeling. Modality alignment~\cite{li2022multi} and dual-stream networks~\cite{chen2022two} further improve CSLR performance by capturing complementary patterns across input types. Multi-stream systems, on the other hand, process the same video through multiple parallel pathways to capture diverse temporal features. AdaBrowse~\cite{hu2023adabrowse} introduces an adaptive selection mechanism to sample informative subsequences and resolutions via a lightweight policy network. Similarly, SlowFastSign~\cite{ahn2024slowfast} extends the SlowFast paradigm developed for action recognition with bidirectional fusion and spatial-temporal enhancement modules, achieving high performance without additional inference overhead. While these methods perform well on CSLR, they typically introduce higher computational cost and require modality-specific supervision, posing limitations for practical deployment.

Most RGB-based CSLR models adopt a pipeline consisting of a spatial encoder followed by a temporal encoder and an alignment module \cite{alyami2024reviewing}. This sequential separation postpones temporal modeling to later stages which prevents early layers from jointly modeling the rich spatio-temporal dynamics crucial for recognizing subtle and context-dependent sign language cues. To address this issue, we propose a Unified Spatial and Temporal Model (USTM) for CSLR. The proposed framework tightly integrates spatial and temporal modeling early in the recognition pipeline. USTM uses a Swin transformer with a custom Temporal Adapter with Positional Embeddings (TAPE) integrated after each feature extraction stage of the proposed model. The TAPE adapter enables joint spatio-temporal feature extraction. This enhanced Swin transformer is followed by a temporal enhancement module that combines a Multi-Scale Temporal Convolution Network (MS-TCN) for capturing fine-grained local dynamics and a bi-directional LSTM to model long-range temporal dependencies. Then, a CTC-based alignment module is used to produce gloss-level outputs without requiring precise temporal annotations.
Our contributions are four-fold: (1) evaluating the effectiveness of transformer-based backbones by comparing ViT and Swin for CSLR; (2) proposing TAPE, a temporal adapter designed to adapt image-based vision backbones for video understanding tasks; (3) conducting an extensive ablation study to investigate the individual contribution of each component within our framework and (4) achieving a new state-of-the-art (SOTA) performance on the PHOENIX14~\cite{phoenix14} and PHOENIX14T~\cite{phoenix14t} datasets, while attaining competitive results on the CSL-Daily~\cite{csldaily} dataset under single-stream RGB-only conditions as well as in multi-modality and multi-stream settings.

The remaining paper is organized as follows: Section~\ref{sec:lit} provides a review of related works; Section~\ref{sec:methodology} describes our proposed framework; Section~\ref{sec:experimental_setup} details our experimental setup, results, and ablation studies; and Section~\ref{sec_conclusion} concludes the paper.
\section{\textbf{Related Work}} \label{sec:lit}

\textbf{Late Stage Temporal Modeling.}  
Most existing CSLR approaches address the recognition problem by first extracting spatial representations independently from individual frames and subsequently performing temporal sequence modeling on the resulting frame-level features. We refer to this paradigm as \emph{late stage temporal modeling}. In this setting, temporal dependencies are modeled on top of compressed spatial representations using sequential modules such as temporal convolutions, recurrent networks, attention mechanisms, or transformers, allowing models to leverage strong image encoders while deferring temporal reasoning to later stages. Within this paradigm, spatial encoding is typically performed using off-the-shelf image backbones. The most commonly adopted backbone is ResNet-18, which serves as the primary spatial encoder in a large portion of late stage temporal modeling approaches~\cite{zhou2021signbert,zheng2023cvt,min2021visual,hu2022temporal,wang2025continuous,yang2025acmc}. Deeper ResNet variants, such as ResNet-34, are also employed in some works to enhance spatial representation capacity~\cite{zhu2025continuous}. Lightweight architectures like MobileNetV2 are explored to reduce computational cost~\cite{ranjbar2025continuous}, while more recent studies adopt transformer-based image encoders, including MViT~\cite{nam2025importance} and Swin Transformers~\cite{alyami2025swin}, to capture hierarchical and multi-scale spatial cues.

Early work adapts image-centric backbones by augmenting them with lightweight temporal mechanisms. Cheng et al.~\cite{cheng2020fully} introduced a fully convolutional framework with gloss feature enhancement to improve alignment. Cui et al.~\cite{cui2019deep} proposed an iterative CNN-RNN training strategy to better capture temporal dependencies. Zhou et al.~\cite{zhou2021signbert} presented a self-emphasizing network that integrates hand-centric cues to improve robustness across signers. Spatial and temporal cues were jointly modeled through multi-cue designs in Zhou et al.~\cite{zhou2021spatial}. Several methods focused on improving temporal aggregation without departing from image backbones. These include multi-scale local temporal similarity fusion~\cite{xie2023multi}, visual alignment constraints~\cite{min2021visual}, self-mutual distillation to reduce CTC spikes~\cite{hao2021self}, and Temporal Lift Pooling for improved temporal downsampling~\cite{hu2022temporal}. Cui et al.~\cite{cui2023spatial} introduced a spatial-temporal transformer that operates on frame chunks. Other extensions explored consistency constraints and auxiliary supervision, such as C2SLR~\cite{zuo2022c2slr}, multi-channel transformers for multi articulatory modeling~\cite{camgoz2020multi}, temporal deformable convolutional fusion of multimodal streams~\cite{papadimitriou2020multimodal}, adaptive temporal browsing in AdaBrowse~\cite{hu2023adabrowse}, and cross-modality augmentation to improve video-text alignment~\cite{pu2020boosting}. Contrastive visual-textual learning with variational alignment was further explored in CVT-SLR~\cite{zheng2023cvt}.

Recent methods strengthen temporal reasoning within modern image transformers and explore hybrid temporal-spatial modules. Alyami et al.~\cite{alyami2025swin} proposed Swin-MSTP, which augments Swin Transformers with multi-scale temporal perception modules for fine-grained motion modeling. Wang et al.~\cite{wang2025continuous} introduced a multi-scale spatial-temporal feature enhancement framework that reinforces frame-wise representations and cross-frame interactions. Huang et al.~\cite{huang2025dual} proposed a dual-stage temporal perception network combining multi-scale local temporal modules with global-local temporal relational modules for richer video features. Tran et al.~\cite{tran2025generalizable} presented a pose-only CSLR framework leveraging local temporal convolutions and region-aware pose encoding for robust generalization. Nam et al.~\cite{nam2025importance} systematically investigated the impact of different facial regions on recognition, showing the mouth is the most critical non-manual feature. Ranjbar et al.~\cite{ranjbar2025continuous} introduced intra-inter gloss attention, exploiting relationships within and between glosses for improved transformer-based CSLR. Yang et al.~\cite{yang2025acmc} proposed ACMC, an adaptive cross-modal multi-grained contrastive learning framework for fine-grained visual-text alignment. Finally, Zhu et al.~\cite{zhu2025continuous} introduced a motor attention mechanism combined with frame-level self-distillation to capture dynamic motion changes, enhancing feature representation and inference robustness.

\textbf{Early Stage Temporal Modeling.}  
In contrast to late stage temporal modeling, early stage temporal modeling refers to CSLR approaches that integrate temporal reasoning directly into the feature extraction process, rather than applying it solely on top of precomputed frame-level representations. In this paradigm, spatial and temporal cues are jointly modeled from the early layers of the network, enabling motion patterns, inter-frame dependencies, and articulatory dynamics to influence feature learning throughout the encoding pipeline. Methods following this approach often rely on spatio-temporal backbones or architectures that explicitly encode temporal information during spatial feature extraction. Representative designs include 3D convolutional networks such as ResNet2+1D-18~\cite{jiang2021skeleton}, S3D~\cite{chen2022two}, and SlowFast-101~\cite{ahn2024slowfast}, which jointly model appearance and motion across frames. Other approaches incorporate temporal interactions through alternative formulations, such as spatio-temporal graph neural networks for structured sign representation~\cite{gan2024signgraph}, attention-based multi-stream keypoint encoders~\cite{guan2405multi}, or by injecting temporal adapters within 2D CNN backbones to enable early temporal coupling~\cite{hu2023continuous}.

Jiang et al.~\cite{jiang2021skeleton} introduced a skeleton-aware multi-modal framework that leverages graph-based skeleton cues alongside RGB and depth inputs. Chen et al.~\cite{chen2022two} proposed TwoStream-SLR, a dual-pathway network that separately encodes raw video frames and keypoint sequences, and uses bidirectional connections, sign pyramid networks, and frame-level self-distillation for enhanced spatio-temporal learning. Hu et al.~\cite{hu2023continuous} presented CorrNet, which dynamically computes correlation maps between adjacent frames to emphasize body trajectories, though the approach may not fully capture long-range temporal dependencies. Gan et al.~\cite{gan2024signgraph} proposed SignGraph, representing sign sequences as graphs to model intra-frame and inter-frame cross-region interactions. Similarly, Ahn et al.~\cite{ahn2024slowfast} developed a SlowFast network with dual pathways for high-temporal and high-spatial resolution processing. Multi-stream keypoint attention models \cite{guan2405multi} and diffusion graph convolution networks with adaptive motion-aware attention \cite{rastgoo2025multi} further explore the combination of skeletal cues, multi-stream fusion, and self-supervised pretraining to capture fine-grained motion and long-term dependencies. Zhou et al.~\cite{zhou2025scaling} introduced large-scale multimodal pretraining, integrating pose and text cues to enhance visual representation learning and contextual understanding. 

These approaches embed temporal reasoning directly within early feature extraction via 3D convolutions, temporal adapters, spatio-temporal graphs, or attention mechanisms, enabling motion-aware representations from the outset. Despite this, early integration of temporal modeling in CSLR remains relatively underexplored.

\section{\textbf{Methodology}} 
\label{sec:methodology}

Given an input video $\mathcal{X} = \{x_1, \dots, x_n\}$ consisting of $n$ frames, the objective of CSLR is to predict a sequence of $m$ glosses $\mathcal{Y} = \{y_1, \dots, y_m\}$ representing the signs that appear in the video. Most existing CSLR frameworks follow a three-stage pipeline consisting of a spatial module $\mathcal{F}_s(\cdot)$ to extract visual features from each frame, a temporal module $\mathcal{F}_t(\cdot)$ to model motion and dependencies across frames, and an alignment module $\mathcal{F}_a(\cdot)$, typically trained with CTC loss, to generate the final gloss sequence. The pipeline can be formulated as follows:
\[
\mathcal{Y} = \mathcal{F}_a(\mathcal{F}_t(\mathcal{F}_s(\mathcal{X})))
\]

\begin{figure*}
    \centering
    \includegraphics[width=2\columnwidth]{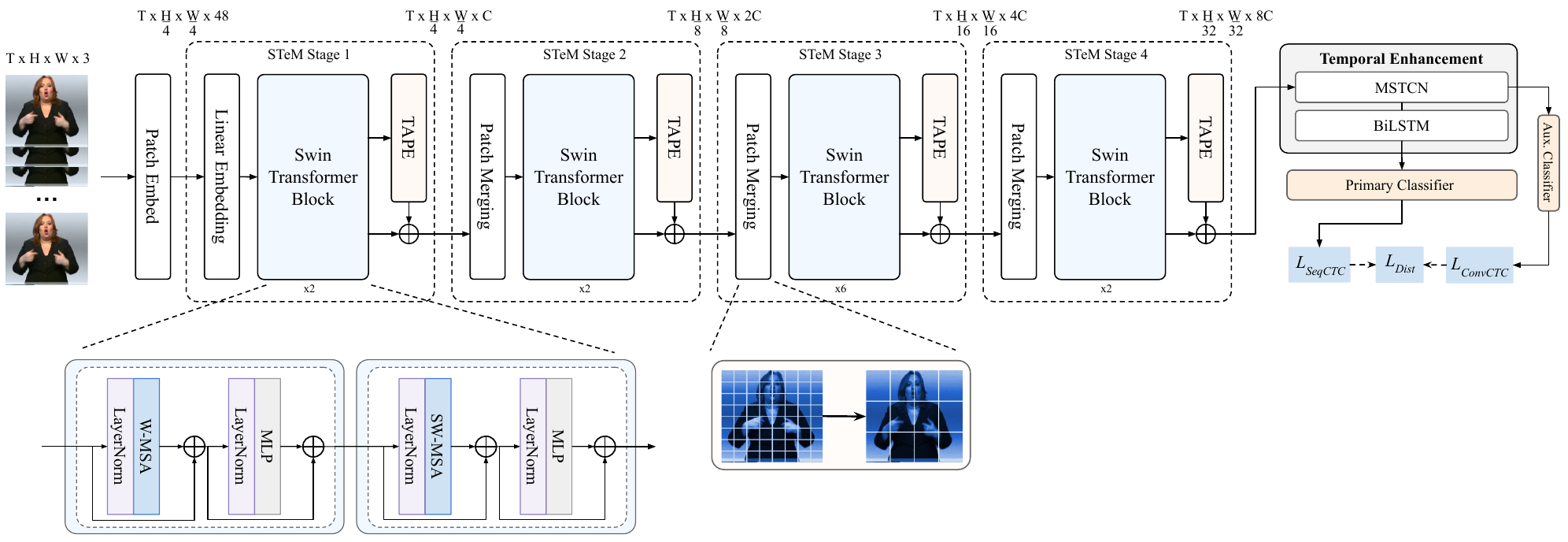}
    \caption{The USTM framework. Spatio-temporal features are extracted using a Swin Transformer backbone with TAPE adapters and subsequently processed by MSTCN and BiLSTM modules in the temporal enhancement stage.}
    \label{fig:pipeline}
\end{figure*}

In contrast to this traditional way of processing spatio-temporal data, we adopt an integrated framework that departs from this strictly sequential design. The proposed framework USTM, consists of a stack of Spatio-Temporal Modules (STeMs) followed by a temporal enhancement stage that refines motion representations using MS-TCN and BiLSTM modules. Finally, a CTC-based alignment module decodes the output into gloss sequences. Figure \ref{fig:pipeline} shows the USTM pipeline. 

\subsection{\textbf{Modeling Spatio-Temporal Features}}
\label{subsec:spatiotemporal}
To effectively model both spatial and temporal patterns in sign language videos, we propose a vision encoder consisting of a stack of STeM blocks. As illustrated in Figure \ref{fig:pipeline}, each STeM block consists of a Swin transformer block for spatial feature extraction, followed by a TAPE adapter for temporal modeling. The motivation for TAPE comes from the observation that temporal modeling in many CSLR architectures typically occurs after spatial features are pooled or flattened into 1D vectors, which limits the ability to capture fine-grained spatio-temporal interactions. TAPE addresses this limitation by applying temporal processing within the spatial feature extraction stage.

\begin{figure}
    \centering
    \includegraphics[width=0.8\columnwidth]{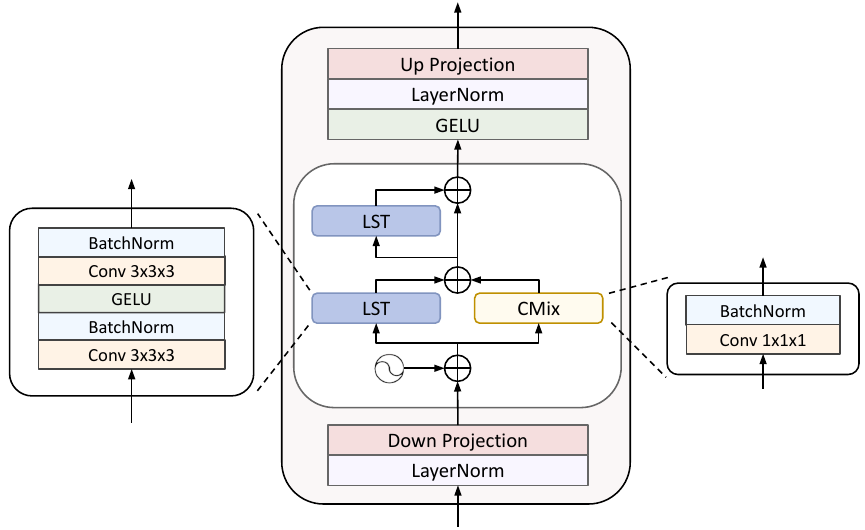}
    \caption{Architecture of TAPE adapter.}
    \label{fig:TAPE}
\end{figure}

The TAPE adapter is a lightweight and pluggable module designed to enhance spatial features with short-term temporal context. As illustrated in Figure \ref{fig:TAPE}, the TAPE adapter begins with layer normalization followed by a down-projection to a lower-dimensional latent space. Learnable temporal positional embeddings are then added to encode frame order. Temporal modeling within TAPE is carried out through two components: Channel Mix (CMix) and Local Spatio-Temporal (LST) modules. The CMix component applies a $1 \times 1 \times 1$ convolution followed by batch normalization, allowing efficient recombination of channel features without expanding the spatio-temporal receptive field. In parallel, the LST component applies two consecutive $3 \times 3 \times 3$ convolutions with GELU activations, each followed by batch normalization, allowing the component to capture fine-grained spatio-temporal dependencies from local spatial and temporal neighborhoods. The outputs from CMix and LST are fused to combine channel-mixed cues with rich local spatio-temporal patterns. A second LST block is then applied to further refine temporal representations. Finally, the fused CMix + LST output is added residually to the output of the second LST block. The combined representation is subsequently passed through GELU activation, layer normalization, and an up-projection layer to restore the original feature dimension.

The TAPE output is then added residually to the output of the preceding Swin Transformer stage, completing one STeM block. The TAPE's hierarchical design allows the network to learn rich spatio-temporal representations efficiently. Furthermore, the TAPE block is architecture-agnostic and can be integrated into other vision backbones for video-based tasks as will be explored further in Section~\ref{sec:ablation}.

\begin{figure}
    \centering
    \includegraphics[width=0.8\columnwidth]{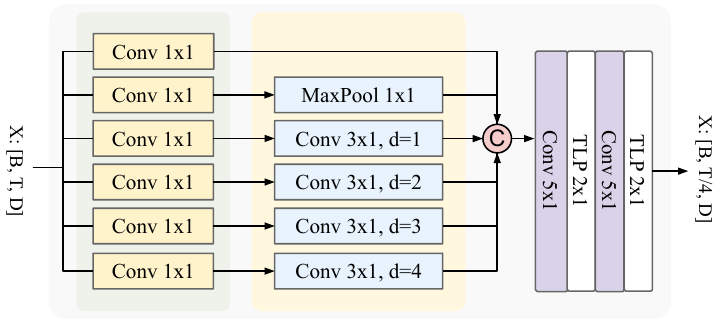}
    \caption{MSTCN architecture with parallel 1D convolutional streams captures multi-scale temporal patterns, followed by 1D convolution, TLP layers, and a BiLSTM for long-term dependencies.}
    \label{fig:MSTCN}
\end{figure}

\subsection{\textbf{Temporal Enhancement with MS-TCN and BiLSTM}}
\label{subsec:temporal}

Following the spatio-temporal encoding stage, we apply a temporal enhancement module to further refine the features with local and global temporal contexts. This module consists of an MS-TCN block followed by a BiLSTM. The MS-TCN is designed to capture temporal information at multiple scales while keeping the computational cost relatively low \cite{alyami2025swin}. It employs several parallel 1D convolutional streams, each configured with a fixed kernel size, $3 \times 1$, and different dilation rates to create diverse receptive fields. In total, the network contains $n + 2$ branches, where $n$ corresponds to the number of dilations and the additional two branches are dedicated to reduce feature dimensionality. In our implementation, we set $n = 4$, resulting in six parallel streams. The outputs of these parallel streams are concatenated and then processed through a sequence of 1D convolution and TLP layers. This structure enables the MS-TCN to effectively capture both fine-grained and coarse-grained temporal patterns, making it suitable for modeling local dependencies. The refined representation is then passed to a BiLSTM layer, which models bidirectional dependencies to capture long-term temporal context. Figure \ref{fig:MSTCN} illustrates the MS-TCN architecture.

\subsection{\textbf{Training Objective}}
\label{subsec:training}

The training objective used in our framework uses a weighted sum of multiple loss functions designed to optimize both early and late stages of the model. We apply a CTC loss after the MS-TCN output ($\mathcal{L}_{\text{ConvCTC}}$) and after the final BiLSTM output ($\mathcal{L}_{\text{SeqCTC}}$). Both loss functions are computed against the ground-truth gloss sequence. To encourage consistency between early and late temporal representations, we include a distillation loss ($\mathcal{L}_{\text{Dist}}$) ~\cite{min2021visual} which aligns the intermediate MS-TCN logits with those produced by the BiLSTM. Additionally, two auxiliary regularization losses from the LiftPool operation within the MS-TCN are incorporated, namely the uncertainty loss ($\mathcal{L}_{\text{Cu}}$) and the peakiness loss ($\mathcal{L}_{\text{Cp}}$), which encourage temporal confidence and sparsity \cite{hu2022temporal}.
The final loss is computed as follows:
\[
\mathcal{L} = \lambda_1 \mathcal{L}_{\text{ConvCTC}} + \lambda_2 \mathcal{L}_{\text{SeqCTC}} + \lambda_3 \mathcal{L}_{\text{Dist}} + \lambda_4 \mathcal{L}_{\text{Cu}} + \lambda_5 \mathcal{L}_{\text{Cp}},
\]
where  $\lambda_1$ and $\lambda_2$ are set to 1.0 to equally supervise both CTC stages. We also assign a higher weight of $\lambda_3 = 25.0$ to the disillation loss to emphasize the importance of distillation, and use smaller values $\lambda_4 = \lambda_5 = 0.001$ for the auxiliary LiftPool losses to act as regularizers without dominating the optimization.
\section{Experiments}
\label{sec:experimental_setup}

\subsection{\textbf{Experimental Setup}}

\textbf{Datasets.} Three benchmarked CSLR datasets have been used in this work to evaluate the proposed framework: PHOENIX14, PHOENIX14T, and CSL-Daily. \textit{PHOENIX14} is a German Sign Language (GSL) dataset compiled from televised weather reports. This dataset features a vocabulary of 1295 unique glosses across 6,841 annotated sentence-level samples, with an average of 11.5 glosses per sentence. The data is performed by nine signers and split into 5,672 training, 629 validation, and 540 test samples. \textit{PHOENIX14T} extends PHOENIX14 with improved gloss annotations and corresponding spoken German translations. This dataset includes 8,247 videos with 1,085 glosses, and an average of 7.7 glosses per sentence. The dataset is split into 7,096 samples for training, 519 for validation, and 642 for testing. \textit{CSL-Daily} is a large-scale Chinese Sign Language dataset comprising 6,588 unique sentences covering approximately 2,000 distinct glosses. Videos are recorded against a uniform white background, and each sentence contains around 7.2 glosses on average. The dataset is divided into 18,401 samples for training, 1,077 for validation, and 1,176 for testing.

\textbf{Implementation Details.} The proposed model was developed using PyTorch and trained on a dual-GPU setup consisting of two NVIDIA RTX A6000 cards, each with 48GB of VRAM. We experimented with three backbone configurations: Swin\textsubscript{tiny}, Swin\textsubscript{small} and Swin\textsubscript{base}. The tiny, small, and base variants differ primarily in model depth, embedding dimensions, and the number of attention heads across stages, with larger variants providing greater capacity at the cost of increased computation. All of these backbones were initialized from ImageNet-1K pretrained checkpoints. 

Each frame in the input video was resized to $256 \times 256$ pixels and random crops of size $224 \times 224$ were extracted during training. At test time, frames were center-cropped to $224 \times 224$. To increase temporal and spatial diversity, the training pipeline included horizontal flipping applied with a probability of 0.5 and temporal jittering within a 20\% range. Training was carried out using an Adam optimizer, with a constant weight decay set to $1 \times 10^{-4}$. For both PHOENIX14 and PHOENIX14T datasets, we trained models for 40 epochs, starting with a learning rate of $10^{-4}$ and decreasing it by a factor of 5 at four milestones: epochs 20, 25, 30, and 35. The CSL-Daily dataset was trained under a different schedule where the learning rate began at $5 \times 10^{-5}$ and was halved at regular intervals, specifically, at epochs 25, 30, 35, 40, and 45 over a total of 50 epochs.

\subsection{\textbf{Comparison with Other Works}} \label{sec:sota}

In this section, we present a comparative evaluation of our proposed model against existing approaches. Specifically, we compare our model with single-stream RGB-based methods, multi-stream frameworks, and multimodal systems. Following prior works \cite{ahn2024slowfast, alyami2025swin, hu2024corrnet+, alyami2025swin}, we employ the Word Error Rate (WER) as our primary evaluation metric. It is computed by calculating the number of insertions, deletions, and substitutions required to transform the predicted sequence into the ground-truth sequence. A lower WER indicates higher recognition accuracy.

\vspace{2mm}\noindent\textbf{RGB-Based Frameworks.} 
Table \ref{tab:sota_singlestream} compares the performance of our USTM framework with prior single-stream RGB-based methods. As shown in the table, USTM variants demonstrate consistent improvements across all datasets. On Phoenix14, USTM\textsubscript{tiny} achieves WERs of 18.1 and 18.3 on the dev and test sets, respectively. USTM\textsubscript{small} achieves 17.9 on dev and 17.6 on test, while USTM\textsubscript{base} improves to 17.4 on dev and 17.7 on test. Our smallest model remains highly competitive, trailing CorrNet+ by only 0.1 on both dev and test. USTM\textsubscript{small} surpasses the previous SOTA CorrNet+ with WER of 18.0 on dev and 18.2 on test, setting a new SOTA on this dataset. Notably, USTM\textsubscript{tiny}, which uses the tiny Swin transformer backbone, outperforms Swin-MSTP that relies on the larger Swin\textsubscript{small} variant which underscores the efficacy of the proposed approach.

\begin{table}[ht]
\centering
\tiny
\caption{Comparison with single stream RGB based models.}
\label{tab:sota_singlestream}
\begin{tabular}{llcccccc}
\toprule
\textbf{Method} & \textbf{Venue} & \multicolumn{2}{c}{\textbf{Phoenix14}} & \multicolumn{2}{c}{\textbf{Phoenix14T}} & \multicolumn{2}{c}{\textbf{CSL-Daily}} \\
& & Dev & Test & Dev & Test & Dev & Test \\
\midrule
FCN \cite{cheng2020fully} & ECCV 2020 & 23.7 & 23.9 & 23.3 & 25.1 & -- & -- \\
mLTSF-Net \cite{xie2023multi} & PR 2023 & 22.9 & 23.0 & -- & -- & -- & -- \\
VAC \cite{min2021visual} & ICCV 2021 & 21.2 & 22.3 & -- & -- & -- & -- \\
MSTNet \cite{xie2023multi} & PR 2023 & 20.3 & 21.4 & -- & -- & -- & -- \\
SEN \cite{hu2023self} & AAAI 2023 & 19.5 & 21.0 & 19.3 & 20.7 & -- & -- \\
SMKD \cite{hao2021self} & ICCV 2021 & 20.8 & 21.0 & 20.8 & 22.4 & -- & -- \\
TLP \cite{hu2022temporal} & ECCV 2022 & 19.7 & 20.8 & 19.4 & 21.2 & -- & -- \\
SSSLR \cite{jang2023self} & ICASSP 2023 & 20.9 & 20.7 & 20.5 & 22.3 & -- & -- \\
CTCA \cite{guo2023distilling} & CVPR 2023 & 19.5 & 20.3 & 19.3 & 20.3 & 31.3 & 29.4 \\
ST-Transformer \cite{cui2023spatial} & CIS 2023 & 19.9 & 19.9 & -- & -- & -- & -- \\
CorrNet \cite{hu2023continuous} & CVPR 2023 & 18.8 & 19.4 & 18.9 & 20.5 & 30.6 & 30.1 \\
CorrNet+ \cite{hu2024corrnet+} & arXiv 2024 & 18.0 & 18.2 & \textbf{17.2} & \underline{19.1} & 28.6 & 28.2 \\
Swin-MSTP \cite{alyami2025swin} & Neurocomp. 2025 & 18.1 & 18.7 & 18.7 & 19.7 & 28.3 & 27.1 \\
CLIP-SLA \cite{alyami2025clip} & CVPR 2025 & 19.7 & 19.3 & 19.8 & 19.4 & \textbf{26.0} & \textbf{25.8} \\
MP-SLR\cite{zhou2025scaling} & TPAMI 2025 & 21.2 & 21.2 & 20.1 & 21.3 & 28.6 & 27.9 \\
STNet\cite{wang2025continuous} & IEEE Acc. 2025 & 19.2 & 19.3 & 19.2 & 19.8 & 28.3 & 27.2 \\
IIGA\cite{ranjbar2025continuous} & MMTA 2025 & 20.1 & 20.4 & - & - & - & - \\
ACMC \cite{yang2025acmc} & IMAVIS 2025 & 17.8 & 18.4 & 17.6 & 18.8 & 28.3 & 27.5 \\
\midrule
USTM\textsubscript{tiny} (Ours) & - & 18.1 & 18.3 & 17.8 & 19.4 & 27.8 & 27.6 \\
USTM\textsubscript{small} (Ours) & - & \underline{17.9} & \textbf{17.6} & \underline{17.4} & 19.3 & \underline{27.7} & \underline{26.4} \\
USTM\textsubscript{base} (Ours) & - & \textbf{17.4} & \underline{17.7} & 17.6 & \textbf{18.9} & 27.7 & 26.5 \\
\bottomrule
\end{tabular}
\end{table}

On Phoenix14T, USTM\textsubscript{tiny} obtained WERs of 17.8 and 19.4 on the dev and test sets, respectively, while USTM\textsubscript{small} reduces the WER to 17.4 on dev and 19.3 on test sets. The best performance is achieved by USTM\textsubscript{base} with 17.6 on dev and 18.9 on test sets, marking the lowest test WER among all approaches. Compared to CorrNet+, our base variant closely matches the dev score and still improves the test score.

On CSL-Daily, USTM\textsubscript{tiny} obtains 27.8 on dev and 27.6 on test sets. USTM\textsubscript{base} is slightly better with 27.7 on dev and 26.5 on test, while USTM\textsubscript{small} achieves the best performance with 27.7 on dev and 26.4 on test sets. Our USTM models significantly outperform CorrNet+ and Swin-MSTP. However, CLIP-SLA performs better on CSL-Daily likely due to its parameter-efficient fine-tuning strategy. By only updating LoRA modules within the pre-trained CLIP encoder, the model avoids overfitting to the training data, whereas USTM trains all parameters of the visual backbone, which may lead to slightly reduced generalization on this dataset.

\vspace{2mm}\noindent\textbf{Multi-Stream Frameworks.} Table \ref{tab:sota_multistream} compares the performance of our USTM framework with prior multi-stream frameworks that use multiple streams to capture varied feature representations. On Phoenix14, USTM\textsubscript{small} surpasses all previous methods, including SlowFastSign, which achieved WERs of 18.0 and 18.3 on the dev and test sets, respectively. Our smaller variants achieve performance on par with SlowFastSign, demonstrating the efficiency of our single-stream design. On Phoenix14T, our models surpass previous methods on the dev split, while trailing the prior best on the test split by a WER of 0.2. On CSL-Daily, USTM\textsubscript{small} trails the previous best method by a margin of 2.2 on the dev split and 1.5 on the test split. The performance gap between our models and SlowFastSign can be attributed to two main factors. First, SlowFastSign is pretrained on the large-scale Kinetics-400 dataset, which provides a strong spatiotemporal prior, while our models are only initialized with ImageNet weights. Second, the SlowFast architecture is explicitly designed to model temporal dynamics at multiple scales throughout the network, which inherently benefits continuous sign language recognition. Despite these differences, our single-stream architecture remains highly competitive and demonstrates strong generalization across datasets without relying on multi-stream inputs or specialized temporal pretraining.

\begin{table}[ht]
\centering
\tiny
\caption{Comparison with multi-stream RGB-based models.}
\label{tab:sota_multistream}
\begin{tabular}{lccccccc}
\toprule
\textbf{Method} & Venue & \multicolumn{2}{c}{\textbf{Phoenix14}} & \multicolumn{2}{c}{\textbf{Phoenix14T}} & \multicolumn{2}{c}{\textbf{CSL-Daily}} \\
& & Dev & Test & Dev & Test & Dev & Test \\
\midrule
CMA \cite{pu2020boosting} & ACM MM 2020 & 23.9 & 24.0 & 24.1 & 24.3 & -- & -- \\
AdaBrowse \cite{hu2023adabrowse} & ACM MM 2023 & 19.6 & 20.7 & 19.5 & 20.6 & 31.2 & 30.7 \\
MSKA \cite{guan2025mska} & PR 2025 & 21.7 & 22.1 & 20.1 & 20.5 & 28.2 & 27.8 \\
SlowFastSign \cite{ahn2024slowfast} & ICASSP 2024 & 18.0 & 18.3 & \underline{17.7} & \textbf{18.7} & \textbf{25.5} & \textbf{24.9} \\
SlowFastSign (Reproduced) \cite{ahn2024slowfast} & ICASSP 2024 & 18.1 & 18.4 & 17.9 & 18.9 & 26.9 & 26.7 \\
\midrule
USTM\textsubscript{tiny} (Ours) & - & 18.1 & 18.3 & 17.8 & 19.4 & 27.8 & 27.6 \\
USTM\textsubscript{small} (Ours) & - & \underline{17.9} & \textbf{17.6} & \underline{17.4} & 19.3 & \underline{27.7} & \underline{26.4} \\
USTM\textsubscript{base} (Ours) & - & \textbf{17.4} & \underline{17.7} & \underline{17.6} & \underline{18.9} & 27.7 & 26.5 \\
\bottomrule
\end{tabular}
\end{table}

\vspace{2mm}\noindent\textbf{Multi-Modal Frameworks.} 
We also compare the performance of the USTM framework with other methods that utilize more than one modality. Table \ref{tab:sota_multimodal} compares the results of our framework with prior multi-modal methods. As shown in the table, most of the multi-modal methods use combine RGB with pose information. On the Phoenix14 dataset, USTM\textsubscript{small} achieves the best results, surpassing all previous works. Notably, even our smallest variant, USTM\textsubscript{tiny}, outperforms TwoStream-SLR the SOTA multi-modal CSLR, achieving 18.1 in dev and 18.3 in test. USTM\textsubscript{small} also performs competitively with the same scores, matching or slightly improving over prior SOTA. For Phoenix14T, USTM improves over TwoStream-SLR with 17.7 in dev and 19.3 in test. On CSL-Daily, USTM\textsubscript{small} attains the lowest WER, surpassing TwoStream-SLR, with 33.1 in dev and 32.0 in test sets.

\begin{table*}[ht]
\centering
\scriptsize
\caption{Comparison with multi-modal frameworks.}
\label{tab:sota_multimodal}
\begin{tabular}{l l ccccccc}
\toprule
\textbf{Method} & \textbf{Venue} & \textbf{Modality} & \multicolumn{2}{c}{\textbf{Phoenix14}} & \multicolumn{2}{c}{\textbf{Phoenix14T}} & \multicolumn{2}{c}{\textbf{CSL-Daily}} \\
& & & Dev & Test & Dev & Test & Dev & Test \\
\midrule
DNF \cite{cui2019deep} & IEEE ToM 2019 & RGB+Op.Flow & 23.1 & 22.8 & -- & -- & 32.8 & 32.4 \\
SignBERT \cite{zhou2021signbert} & IEEE Acc. 2021 & RGB+Pose & 20.1 & 20.2 & -- & -- & 33.6 & 33.1 \\
STMC \cite{zhou2021spatial} & IEEE ToM 2021 & RGB+Pose & 20.6 & 21.5 & 19.6 & 21.0 & -- & -- \\
C2SLR \cite{zuo2022c2slr} & CVPR 2022 & RGB+Pose & 20.5 & 20.4 & 20.2 & 20.4 & -- & -- \\
MSTCSLRN \cite{li2022multi} & arXiv 2022 & RGB+Gloss & -- & 22.8 & -- & -- & -- & -- \\
CVT-SLR \cite{zheng2023cvt} & CVPR 2023 & RGB+Pose & 19.8 & 20.1 & 19.4 & 20.3 & -- & -- \\
TwoStream-SLR \cite{chen2022two} & NeurIPS 2022 & RGB+Pose & 18.4 & 18.8 & 17.7 & 19.3 & 33.1 & 32.0 \\
\midrule
USTM\textsubscript{tiny} (Ours) & - & RGB & 18.1 & 18.3 & 17.8 & 19.4 & 27.8 & 27.6 \\
USTM\textsubscript{small}& - & RGB & \underline{17.9} & \textbf{17.6} & \underline{17.4} & 19.3 & \underline{27.7} & \underline{26.4} \\
USTM\textsubscript{base} (Ours) & - & RGB & \textbf{17.4} & \underline{17.7} & \textbf{17.6} & \textbf{18.9} & \underline{27.7} & \underline{26.5} \\
\bottomrule
\end{tabular}
\end{table*}

\subsection{\textbf{Ablation Studies}} \label{sec:ablation}

\begin{figure*}
    \centering
    \includegraphics[width=2\columnwidth]{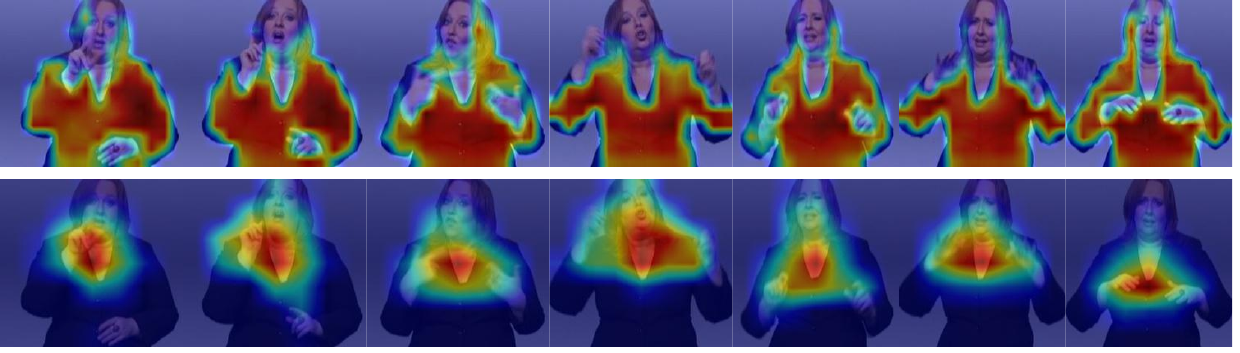}
    \caption{\textbf{EigenCAM visualization of a sample from the Phoenix2014 dataset.} The top row illustrates the EigenCAM of the USTM with a ViT backbone, while the bottom row illustrates the EigenCAM of the USTM with a Swin transformer backbone. The red color indicates the most attended regions.}
    \label{fig:EigenCAM}
\end{figure*}

To analyze the contribution of our framework's components, we conducted a series of ablation studies on the Phoenix2014 dataset. These experiments evaluate the effect of the temporal adapter design, the spatio-temporal encoding strategy, and core architectural hyperparameters. In all experiments involving temporal adapters, we follow a consistent placement strategy unless stated otherwise. Specifically, for the Swin transformer backbone, the adapters are inserted after each of the four Swin transformer stages, whereas for the ViT backbone, they are integrated after every three transformer blocks.

\vspace{2mm}\noindent\textbf{Temporal Adapter Design.}  
The TAPE adapter lies at the core of our USTM pipeline. We first evaluate whether incorporating adapters within the spatial backbone is beneficial to the overall performance of the proposed framework. We adopt STAdapter \cite{pan2022st} as a baseline spatiotemporal adapter and compare it with the proposed TAPE adapter. In the temporal enhancement stage, we employed a simple Conv1D + BiLSTM module instead of MS-TCN + BiLSTM, ensuring that the effect of MS-TCN does not confound the comparison. As shown in Table~\ref{tab:adapter_pe}, interleaving STAdapter blocks within the spatial backbone substantially improves performance. The effect is most apparent with the ViT backbone, where WER drops from 43.2 to 22.3 on dev and from 43.3 to 22.1 on test. A similar trend is observed with the Swin transformer backbone, where WER decreases by 0.6 on dev and 0.8 on test. Replacing STAdapter with TAPE yields further improvements, reducing WER on ViT to 20.2 and 20.7, and on Swin\textsubscript{small} to 18.4 and 19.1 on dev and test, respectively. 

\begin{table}[ht]
\centering
\scriptsize
\caption{Effect of spatiotemporal adapters.}
\label{tab:adapter_pe}
\begin{tabular}{llcc}
\toprule
\textbf{Backbone} & \textbf{Adapter} & \textbf{Dev} & \textbf{Test} \\
\midrule
ViT & - & 43.2 & 43.3 \\
ViT & STAdapter & 22.3 & 22.1 \\
ViT & TAPE & 20.2 & 20.7\\
Swin\textsubscript{small} & - & 19.9 & 19.9 \\
Swin\textsubscript{small} & STAdapter & 19.3 & 19.1 \\
Swin\textsubscript{small} & TAPE & \textbf{18.4} & \textbf{19.1} \\
\bottomrule
\end{tabular}
\end{table}

\vspace{2mm}\noindent\textbf{Temporal Adapter Insertion Frequency.}  
For deeper architectures, such as ViT-B16, which consists of 12 layers compared to only 4 stages in Swin, we studied how frequently TAPE adapters should be inserted. To control this, we introduce a hyperparameter $\alpha$ that defines the interval between two TAPE adapters, where smaller $\alpha$ values correspond to denser insertion. As shown in Table~\ref{tab:alpha_ablation}, inserting TAPE after every single layer ($\alpha = 1$) or too sparsely ($\alpha = 3$) results in weaker performance. The best performance is achieved with $\alpha = 2$, suggesting that moderate temporal injection offers the right balance between model capacity and temporal granularity, allowing effective exploitation of local temporal structure without oversaturating the backbone.

\begin{table}[ht]
\centering
\scriptsize
\caption{Effect of varying adapter insertion interval $\alpha$ in ViT-B16.}
\label{tab:alpha_ablation}
\begin{tabular}{ccc}
\toprule
\textbf{Adapter Interval $\alpha$} & \textbf{Dev} & \textbf{Test} \\
\midrule
1 & 22.8 & 22.6 \\
2 & \textbf{20.4} & \textbf{20.0} \\
3 & 20.8 & 20.7 \\
\bottomrule
\end{tabular}
\end{table}

\vspace{1mm}\noindent\textbf{USTM Components.}  
We first analyze the choice of backbone to study its impact on the USTM pipeline. ViT-B16 and Swin\textsubscript{small} were compared under minimal configurations using only a simple Conv1D and BiLSTM in the temporal enhancement stage. As shown in Table~\ref{tab:component_ablation}, ViT-B16 performs poorly with a WER of 43.2 on dev and 43.3 on test, while Swin\textsubscript{small} achieves a WER of 19.8 on dev and 19.9 on test. The superiority of Swin over ViT is further evident in the EigenCam \cite{muhammad2020eigen} visualizations in Figure~\ref{fig:EigenCAM}. The Swin-based model focuses clearly on key regions, such as the hands and mouth, producing smooth and semantically meaningful attention maps. In contrast, the ViT-based model attends to irrelevant areas like the torso, with noisy and fragmented attention. This highlights Swin’s strength in capturing spatially dense cues through its hierarchical and windowed attention, reinforcing its effectiveness for CSLR tasks.

\begin{table}[ht]
\centering
\scriptsize
\caption{Ablation study of USTM core components, evaluating the effects of backbone selection, TAPE, and MS-TCN.}

\label{tab:component_ablation}
\begin{tabular}{llcccc}
\toprule
\multicolumn{2}{c}{\textbf{Backbone}} & \textbf{TAPE} & \textbf{MS-TCN} & \textbf{Dev} & \textbf{Test} \\
\cmidrule(lr){1-2}
\textbf{ViT-B16} & \textbf{Swin\textsubscript{small}} & & & & \\
\midrule
\checkmark & & & & 43.2 & 43.3 \\
\checkmark & & & \checkmark & 19.3 & 19.6 \\
\checkmark & & \checkmark & & 20.2 & 20.7 \\
\checkmark & & \checkmark & \checkmark & 20.8 & 20.7 \\
\midrule
& \checkmark & & & 19.8 & 19.9 \\
& \checkmark & & \checkmark & 18.1 & 18.7 \\
& \checkmark & \checkmark & & 18.4 & 19.1 \\
& \checkmark & \checkmark & \checkmark & \textbf{17.9} & \textbf{17.6} \\
\bottomrule
\end{tabular}
\end{table}

Next, we investigate the interplay between the TAPE adapter and MS-TCN as temporal modeling components. As previously established, TAPE alone improves performance on both ViT-B16 and Swin transformer backbones. Incorporating MS-TCN further refines temporal modeling. For ViT-B16, MS-TCN alone reduces WER to 19.3 and 19.6 on dev and test respectively. However, MS-TCN combined with TAPE increase the WER, indicating some redundancy in this configuration. For Swin\textsubscript{small}, however, combining MS-TCN with TAPE yields the best overall performance of 17.9 on dev and 17.6 on test. These findings indicate that MS-TCN and TAPE interact differently across backbones, showing complementary behavior for Swin but partial redundancy for ViT-B16. This suggests that their combined benefit is architecture-dependent.

\section{Conclusion}
\label{sec_conclusion}
In this work, we introduced USTM, a unified spatio-temporal modeling framework for CSLR. By integrating hierarchical visual backbones, such as the Swin Transformer, with the TAPE adapter and MS-TCN temporal modules, USTM effectively captures fine-grained spatial details and multi-scale temporal dependencies in sign language videos. Our design allows temporal information to be directly injected into the visual backbone while simultaneously modeling long-range motion patterns, resulting in robust and semantically meaningful representations. Extensive experiments on PHOENIX14, PHOENIX14T, and CSL-Daily datasets demonstrate that USTM consistently outperforms prior single-stream, multi-stream, and multi-modal CSLR approaches, establishing new SOTA performance. Ablation studies further confirm the importance of adapter placement, temporal modeling, and backbone selection, while EigenCAM quantitative analysis illustrated the model’s capability in capturing critical regions on the signer, such as the hands and face. Despite its strong performance, USTM relies on a relatively large Swin transformer backbone, which can pose computational challenges for real-time deployment. Future work may explore model compression, efficient temporal adapters, and multi-modal extensions to enhance generalization and enable deployment in resource-constrained environments.

\section*{Acknowledgments}
The authors would like to acknowledge the support provided by King Fahd University of Petroleum \& Minerals (KFUPM) for funding this work through project number ISP24226.
\bibliography{references}
\bibliographystyle{unsrt}

\end{document}